\documentclass[]{article}
\usepackage{geometry}
\usepackage[backend=biber]{biblatex}
\usepackage{authblk}
\usepackage{graphicx}
\usepackage{epstopdf}
\usepackage{multirow}
\usepackage{booktabs}
\usepackage{algpseudocode,algorithm,algorithmicx}
\usepackage{amsmath}
\usepackage{amssymb}
\usepackage{bbm}
\usepackage{lmodern}
\usepackage{titlesec}
\usepackage{hyperref}
\usepackage{dblfloatfix}
\usepackage{subcaption}

\titleformat{\subsubsection}[runin]
  {\normalfont\normalsize\bfseries}{\thesubsubsection}{1em}{}
\titlespacing*{\subsubsection}{0pt}{0.7ex}{1.5ex plus .2ex}

\title{DeepMag: Source Specific Motion Magnification \\Using Gradient Ascent}
\author[1]{Weixuan Chen}
\author[2]{Daniel McDuff}

\affil[1]{Massachusetts Institute of Technology, Cambridge, Massachusetts}
\affil[2]{Microsoft Research, Redmond, Washington}

\date{June 2018}

\begin{document}


\maketitle

\begin{abstract}
Many important physical phenomena involve subtle signals that are difficult to observe with the unaided eye, yet visualizing them can be very informative. Current motion magnification techniques can reveal these small temporal variations in video, but require precise prior knowledge about the target signal, and cannot deal with interference motions at a similar frequency. We present DeepMag an end-to-end deep neural video-processing framework based on gradient ascent that enables automated magnification of subtle color and motion signals from a specific source, even in the presence of large motions of various velocities. While the approach is generalizable, the advantages of DeepMag are highlighted via the task of video-based physiological visualization. Through systematic quantitative and qualitative evaluation of the approach on videos with different levels of head motion, we compare the magnification of pulse and respiration to existing state-of-the-art methods. Our method produces magnified videos with substantially fewer artifacts and blurring whilst magnifying the physiological changes by a similar degree.
\end{abstract}

%
%
%
%
\renewcommand*{\thefootnote}{\arabic{footnote}}

\begin{figure*}[t]
	  \centering\noindent
	  \includegraphics[width=\textwidth]{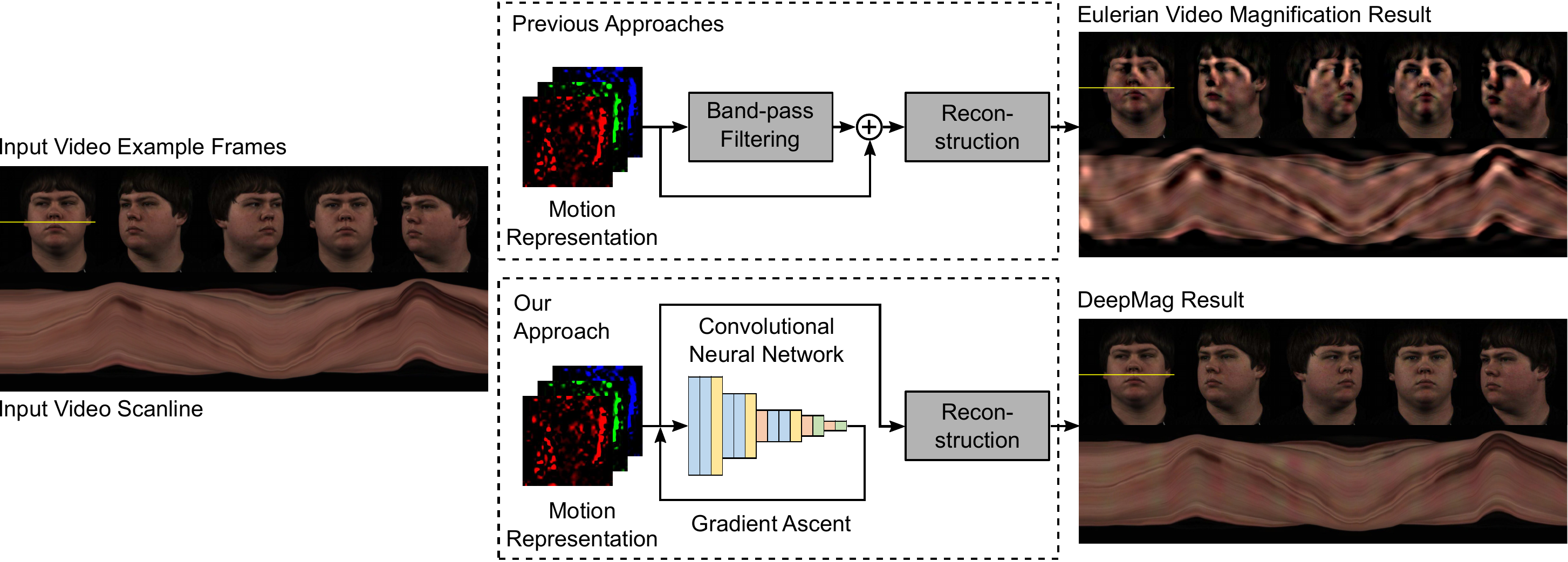}
	  \caption{We present a novel end-to-end deep neural framework for video magnification. Our method allows measurement, magnification and synthesis of subtle color and motion changes from a specific source even in the presence of large motions. We demonstrate this via pulse and respiration manipulation in 2D videos. Our approach produces magnified videos with substantially fewer artifacts when compared to the state-of-the-art.}
	  \label{fig:overview}
\end{figure*}

\section{Introduction}
Revealing subtle signals in our everyday world is important for helping us understand the processes that cause them. Magnifying small temporal variations in video has applications in both basic science (e.g., visualizing physical processes in the world), engineering (e.g., identifying the motion of large structures) and education (e.g, teaching scientific principals). To provide an illustration, physiological phenomena are often invisible to the unaided eye, yet understanding these processes can help us detect and treat negative health conditions. Pulse and respiration magnification specifically, are good exemplar tasks for video magnification as physiological phenomena cause both subtle color and motion variations. Furthermore, larger rigid and non-rigid motions of the body often mask the subtle variations, which makes the magnification of physiological signals non-trivial. 

Several methods have been proposed to reveal subtle temporal variations in video. \textit{Lagrangian} methods for video magnification~\cite{liu2005motion} rely on accurate tracking of the motion of particles (e.g., via optical flow) over time. These approaches are computationally expensive and will not work effectively for color changes. \textit{Eulerian} video magnification methods do not rely on motion estimation, but rather magnify the variation of pixel values over time~\cite{wu2012eulerian}. This simple and clever approach allows for subtle signals to be magnified that might otherwise be missed by optical flow. Subsequent iterations of such approaches have improved the method with phase-based representations \cite{wadhwa2013phase}, matting \cite{elgharib2015video}, second-order manipulation \cite{Zhang2017}, and learning-based representations \cite{oh2018learning}. However, all these approaches use frequency properties to separate the target signal from noise, so they require precise prior knowledge about the signal frequency. Furthermore, if the signal of interest is at a similar frequency to another signal (for example if head motions are at a similar frequency as the pulse signal) an Eulerian approach will magnify both and cause numerous artifacts (see Fig. \ref{fig:overview}).  

To address these problems, we present a generalized approach for magnifying color and motion variations in videos that feature other periodic or random motions. Our method leverages a convolutional neural network (CNN) as a video motion discriminator to separate a specific source signal even if it overlaps with other motion sources in the frequency domain.  Then the separated signal can be magnified in video by performing gradient ascent~\cite{erhan2009visualizing} in the input space of the CNN, with the other motion sources untouched. To adapt the gradient ascent method to the video magnification task, several methodological innovations are introduced including adding L1 normalization and sign correction. The whole algorithm proves to work effectively even in the presence of interference motions with large magnitudes and velocities. Fig. \ref{fig:overview} shows a comparison between the proposed method and previous approaches.


While our method can generally be applied to any type of color or motion magnification task, magnifying physiological changes on the human body without impacting other aspects of the visual appearance is an especially interesting use case with numerous applications in and of itself. In medicine and affective computing the photoplethysmogram (PPG) and respiration signals are used as unobtrusive measures of cardiopulmonary performance. Visualizing these signals could help in the understanding vascular disease, heart conditions (e.g., arterial fibrillation) \cite{chan2016diagnostic} and stress responses. For example, jugular venous pressure (JVP) is analyzed by studying subtle motions of the neck. This is challenging for clinicians and video-magnification could offer a practical aid. Another application is in the design of avatars~\cite{suwajanakorn2017synthesizing}. Synthetic embodied agents may fall into the ``uncanny valley"~\cite{mori1970uncanny} or be easily detected as ``spoofs" if they do not exhibit accurate physiological responses, including respiration, pulse rates and blood flow that can be recovered using video analysis~\cite{poh2010non}. Our method presents the opportunity to not only magnify signals but also synthesize them at different frequencies within a video.

The main contributions of this paper are to: (1) present our novel end-to-end framework for video magnification based on a deep convolutional neural network and gradient ascent, (2) demonstrate recovery of the pulse and respiration waves and magnification of these signals in the presence of large rigid head motions, (3) systematically quantitatively and qualitatively compare our approach with state-of-the-art motion magnification approaches under different rigid motion conditions.

\section{Related Work}

\subsection{Video Motion Magnification}
Lagrangian video magnification approaches involve estimation of motion trajectories that are then amplified~\cite{liu2005motion,wang2006cartoon}. However, these approaches require a number of complex steps including, performing a robust registration, frame intensity normalization, tracking and clustering of feature point trajectories, segmentation and magnification. Another approach, using temporal sampling kernels can aid visualization of time-varying effects within videos~\cite{fuchs2010real}. However, this method involves video downsampling and relies on high framerate input videos.

The neat Eulerian video magnification (EVM) approach proposed by Wu et al.~\cite{wu2012eulerian} combines spatial decomposition with temporal filtering to reveal time varying signals without estimating motion trajectories. However, it uses linear magnification that only allows for relatively small magnifications at high spatial frequencies and cannot handle spatially variant magnification. To counter the limitation, Wadhwa et al.~\cite{wadhwa2013phase} proposed a non-linear phase-based approach, magnifying phase variations of a complex steerable pyramid over time. Replacing the complex steerable pyramid~\cite{wadhwa2013phase} with a Riesz pyramid~\cite{wadhwa2014riesz} produces faster results. In general, the linear EVM technique is better at magnifying small color changes, while the phase-based pipeline is better at magnifying subtle motions~\cite{Wu2012web}. Both the EVM and the phase-EVM techniques rely on hand-crafted motion representations. To optimize the representation construction process, a learning-based method \cite{oh2018learning} was proposed, which uses convolutional neural networks as both frame encoders and decoders. With the learned motion representation, fewer ringing artifacts and better noise characteristics have been achieved.

One common problem with all the methods above is that they are limited to stationary objects, whereas many realistic applications would involve small motions of interest in the presence of large ones. After motion magnification, these large motions would result in large artifacts such as haloes or ripples, and overwhelm any small temporal variation. A couple of improvements have been proposed including a clever layer-based approach called DVMAG \cite{elgharib2015video}. By using matting, it can amplify only a specific region of interest (ROI) while maintaining the quality of nearby regions of the image. However, the approach relies on 2D warping (either affine or translation-only) to discount large motions, so it is only good at diminishing the impact of motions parallel to the camera plane and cannot deal with more complex 3D motions such as the human head rotation. The other method addressing large motion interferences is video acceleration magnification (VAM) \cite{Zhang2017}. It assumes large motions to be linear on the temporal scale so that magnifying the motion acceleration via a second-order derivative filter will only affect small non-linear motions. However, the method will fail if the large motions have any non-linear components, and ideal linear motions are rare in real life, especially on living organisms.

Another problem with all the previous motion magnification methods is that they use frequency properties to separate target signals from noise, so they typically require the frequency of interest to be known a priori for the best results and, as such, have at least three parameters (the frequency bounds and a magnification factor) that need to be tuned. If there are motion signals from different sources that are at similar frequencies (e.g., someone is breathing and turning their head), it is previously not possible to isolate the different signals.


\subsection{Gradient Ascent for Feature Visualization}
Opposite to gradient descent, gradient ascent is a first-order iterative optimization algorithm that takes steps proportional to the positive of the gradient (or approximate gradient) of a function. Since neural networks are generally differentiable with respect to their inputs, it is possible to perform gradient ascent in the input space by freezing the network weights and iteratively tweaking the inputs towards the maximization of an internal neuron firing or the final output behavior. Early works found that this technique can be used to visualize network features (showing what a network is looking for by generating examples)~\cite{erhan2009visualizing,simonyan2013deep} and to produce saliency maps (showing what part of an example is responsible for the network activating a particular way)~\cite{simonyan2013deep}.

A recent famous application of gradient ascent in feature visualization is Google DeepDream~\cite{mordvintsev2015deepdream}. It maximizes the L2 norm of activations of a particular layer in a CNN to enhance patterns in images and create a dream-like hallucinogenic appearance. It should be noted that applying gradient ascent independently to each pixel of the inputs commonly produces images with nonsensical high-frequency noise, which can be improved by including a regularizer that prefers inputs that have natural image statistics. Also, following the same idea of DeepDream, not only a network layer but also a single neuron, a channel, or an output class can be set as the objective of gradient ascent. For a comprehensive discussion of various regularizers and different optimization objectives used in feature visualization tasks see ~\cite{olahfeature}. 

None of the previous works have applied gradient ascent to motion magnification or any task related to motions in video. In contrast to DeepDream and similar visualization tools, our method maximizes the output activation of a CNN in motion representations computed from frames instead of in raw images.

\subsection{Video-Based Physiological Measurement}
Over the past decade video-based physiological measurement using RGB cameras has developed significantly~\cite{mcduff2015survey}. For instance, physiological parameters such as heart rate (HR) and breathing rate (BR) have been accurately extracted from facial videos in which subtle color changes of the skin caused by blood circulation can be amplified and analyzed (a.k.a., imaging plethysmography)~\cite{verkruysse2008remote,poh2010non,poh2011advancements,de2013robust,Tarassenko2014,Wang2016b}. Similar metrics have also been extracted by analyzing subtle face motions associated with the blood ejection into the vessels (a.k.a., imaging ballistocardiography)~\cite{Balakrishnan2013} as well as more prominent chest volume changes during breathing~\cite{Tan2010,Janssen2016}.

Early work on imaging plethysmography identified that spatial averaging of skin pixel values from an imager could be used to recover the blood volume pulse~\cite{takano2007heart}. The strongest pulse signal was observed in the green channel~\cite{verkruysse2008remote}, but a combination of color channels provides improved results~\cite{poh2010non,mcduff2014improvements}. Combining these insights with face tracking and signal decomposition enables a fully automated recovery of the pulse wave and heart rate~\cite{poh2010non}.

In the presence of dynamic lighting and motion, advancements were needed to successfully recover the pulse signal. Leveraging models grounded in the optical properties of the skin has improved performance. The CHROM~\cite{de2013robust} method uses a linear combination of the chrominance signals. It makes the assumption of a standardized skin color profile to white-balance the video frames. The Pulse Blood Vector (PBV) method~\cite{de2014improved} relies on characteristic blood volume changes in different regions of the frequency spectrum to weight the color channels. Adapting the facial ROI can improve the performance of iPPG measurements as blood perfusion varies in intensity across the body~\cite{Tulyakov2016}

Few approaches have made use of supervised learning for video-based physiological measurement.  Formulating the problem is not trivial and performance has been modest~\cite{osman2015supervised,monkaresi2014machine}. Recent advances in deep neural video analysis offer opportunities for recovering accurate physiological measurements. Recently, Chen and McDuff~\cite{chen2018deepphys} presented a supervised method using a convolutional attention network that provided state-of-the-art measurement performance and generalized across people. Our video magnification algorithm is based on a novel framework that allows recovery of pulse and respiratory waves using such a convolutional architecture. 

\section{Methods}

\begin{figure*}
	\centering\noindent
	\includegraphics[width=\linewidth]{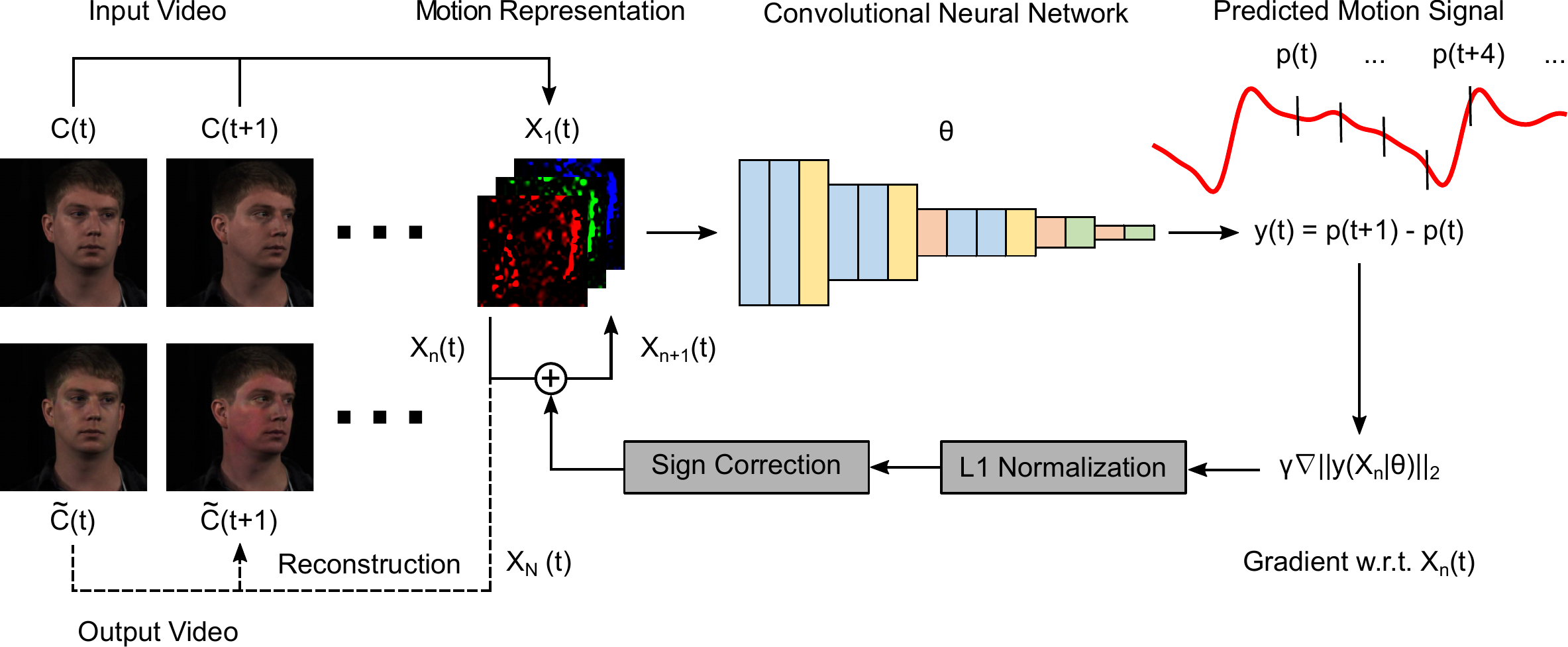}
	\caption{The architecture of DeepMag. The CNN model predicts the motion signal of interest based on a motion representation computed from consecutive video frames. Magnification of the motion signal in video can be achieved by amplifying the L2 norm of its first-order derivative and then propagating the changes back to the motion representation using gradient ascent.}  
	\label{fig:architechture}
\end{figure*}

\subsection{Video Magnification Using Gradient Ascent}

Fig. \ref{fig:architechture} shows the workflow of the proposed video magnification algorithm using gradient ascent. Similar to previous video magnification algorithms, it reads a series of video frames $C(t),~t=1,2,\cdots,T$, magnifies a specific subtle motion in them, and outputs frames of the same dimension $\widetilde{C}(t),~t=1,2,\cdots,T$.  

The first step of our algorithm is computing the input motion representation $X_1(t)$ from the original video frames $C(t),~t=1,2,\cdots,T$. $X_1(t)$ represents any change happening between two consecutive frames $C(t)$ and  $C(t+1)$. Common motion representations include frame difference and optical flow. Different motion representations can emphasize different aspects of motions. For example, the physio-logy-based motion representation called normalized frame difference \cite{chen2018deepphys} was proposed to capture skin absorption changes robustly under varying rigid motions. On the other hand, optical flow based on the brightness constancy constraint is good at representing object displacements, but largely ignores the light absorption changes of objects. As a general framework for video magnification, our algorithm supports any type of motion representation.

In realistic videos the motion representations are comprised of multiple motions from different sources. For example, unconstrained facial video recordings commonly contain not only respiration movements and pulse-induced skin color changes but also head rotations and facial expressions. As we are only interested in magnifying one of these motions at a time, a video magnification algorithm should have the ability to separate the target motion from the others in the motion representation. Previous methods have typically used frequency-domain characteristics of the target motion in separation, so they rely on precise prior knowledge about the motion frequency (e.g. the exact heart rate). Furthermore, if any other motion overlaps with the target motion in frequency, it will still be magnified and cause artifacts. To improve the specificity of magnification and reduce the dependence on prior knowledge, we propose to use a deep convolutional neural network (CNN) to model the relationship between the motion representation and the motion of interest. As shown in Fig. \ref{fig:architechture}, the CNN has the input motion representation $X_1(t)$ as its input, and the first-order derivative $y(t)$ of the target motion signal $p(t)$ as its output. For many motion types, there are available datasets with paired videos and ground truth motion signals (e.g., facial videos with pulse and respiration signals measured from medical devices). Therefore, the weights $\theta$ of the CNN can be determined by training it on one of these datasets. It has been shown in \cite{chen2018deepphys} that CNNs trained in this way have good generalization ability over different objects (human subjects), different backgrounds, and different lighting conditions. 

As the CNN has established the relationship between the input motion representation $X_1(t)$ and the target motion signal $p(t)$, magnification of $p(t)$ in $X_1(t)$ can be achieved by amplifying the L2 norm of its first-order derivative $y(t)$ and then propagating the changes back to $X_1(t)$ using gradient ascent. The process can be expressed as 
\begin{equation} \label{eq:I}
X_{n+1} = X_{n}+\gamma\nabla\|y(X_{n}|\theta)\|_2,~n=1,2,\cdots,N-1
\end{equation}
in which $N$ is the total number of iterations and $\gamma$ is the step size. $\theta$ is the weights of the CNN, which are frozen during gradient ascent. $\nabla\|y(X_{n}|\theta)\|_2$ is the gradient of $\|y(t)\|_2$ with respect to $X_n(t)$, which is the direction to which $X_n(t)$ can be modified to specifically magnify the target motion rather than the other motions. Note that both $X_{n}$ and $y$ correspond to time point $t$ in (\ref{eq:I}), but $t$ is omitted for conciseness.

The vanilla gradient ascent in (\ref{eq:I}) is appropriate for magnifying a single motion representation $X_1(t)$ at time $t$. However, for video magnification, a series of motion representations $X_1(t) ,~t=1,2,\cdots,T$ need to be processed and magnified to the same level. Since the magnitude of the gradient is sensitive to the surface shape of the objective function (i.e. a point on a steep surface will have high magnitude whereas a point on the fairly flat surface will have low magnitude), it is not guaranteed that the accumulated gradient will be proportional to the original motion amplitude. Therefore, we apply L1 normalization to the gradient
\begin{equation} \label{eq:II}
X_{n+1} = X_{n}+\gamma\frac{\nabla\|y(X_{n}|\theta)\|_2}{\|\nabla\|y(X_{n}|\theta)\|_2\|_1}
\end{equation}
so that only the gradient direction is kept and the gradient magnitude is controlled by the step size $\gamma$.

Another problem with (\ref{eq:I}) is that motions in opposite directions contribute equivalently to the L2 norm of $y(t)$. As a result, the target motion might be amplified in terms of the absolute amplitude but 180-degrees out of phase. To address the problem, we correct the signs of the gradient to always match the signs of the input motion representation \begin{equation} \label{eq:III}
X_{n+1} = X_{n}+\gamma\frac{\nabla\|y(X_{n}|\theta)\|_2\odot sgn(X_{n}\odot\nabla\|y(X_{n}|\theta)\|_2)}{\|\nabla\|y(X_{n}|\theta)\|_2\|_1}
\end{equation}
in which $sgn(\cdot)$ is the sign function and $\odot$ is element-wise multiplication.

Summing up the changes of $X_n(t)$ in all the iterations, we get the final expression of the magnified motion representation:
\begin{equation} \label{eq:IV}
X_N = X_1+\sum_{n=1}^{N-1}{\gamma\frac{\nabla\|y(X_{n}|\theta)\|_2\odot sgn(X_{n}\odot\nabla\|y(X_{n}|\theta)\|_2)}{\|\nabla\|y(X_{n}|\theta)\|_2\|_1}}
\end{equation}
There are only two hyper-parameters $\gamma$ and $N$, which can be tuned to change the magnification factor. Finally, the magnified motion representation can be combined with previous frames to iteratively generate the output video. The complete algorithm is summarized in Algorithm \ref{alg1}.

\begin{algorithm}
	\caption{DeepMag video magnification
		\label{alg1}}
	\begin{algorithmic}[1]
		\Require $C(t),~t=1,2,\cdots,T$ is a series of video frames, $\mathcal{M}$ is a motion representation estimator, $\theta$ is the pre-trained CNN weights for predicting a target motion signal $y$, $\gamma$ is the step size, and $N$ is the number of iterations
		\For {$t = 1 \textrm{ to } T-1$}
		\State Compute motion representation: $X_1(t)\gets \mathcal{M}(C(t),C(t+1))$
		\For {$n = 1 \textrm{ to } N-1$}
		\State Compute gradient: $G_n(t)\gets \nabla\|y(X_{n}(t)|\theta,t)\|_2$
		\State L1 normalization: $G_n(t)\gets G_n(t)/\|G_n(t)\|_1$
		\State Sign correction: $G_n(t)\gets G_n(t)\odot sgn(G_n(t)\odot X_n(t))$
		\State Gradient ascent: $X_{n+1}(t)\gets X_n(t)+\gamma G_n(t)$
		\EndFor
		\EndFor
		\State $\widetilde{C}(1)=C(1)$
		\For {$t = 1 \textrm{ to } T-1$}
		\State Reconstruct magnified frame $\widetilde{C}(t+1)\gets\mathcal{M}^{-1}(\widetilde{C}(t),X_N(t))$
		\EndFor
		\State \Return{$\widetilde{C}(t),~t=1,2,\cdots,T$}
	\end{algorithmic}
\end{algorithm}

\begin{figure}[t]
	\centering\noindent
	\includegraphics[width=0.5\linewidth]{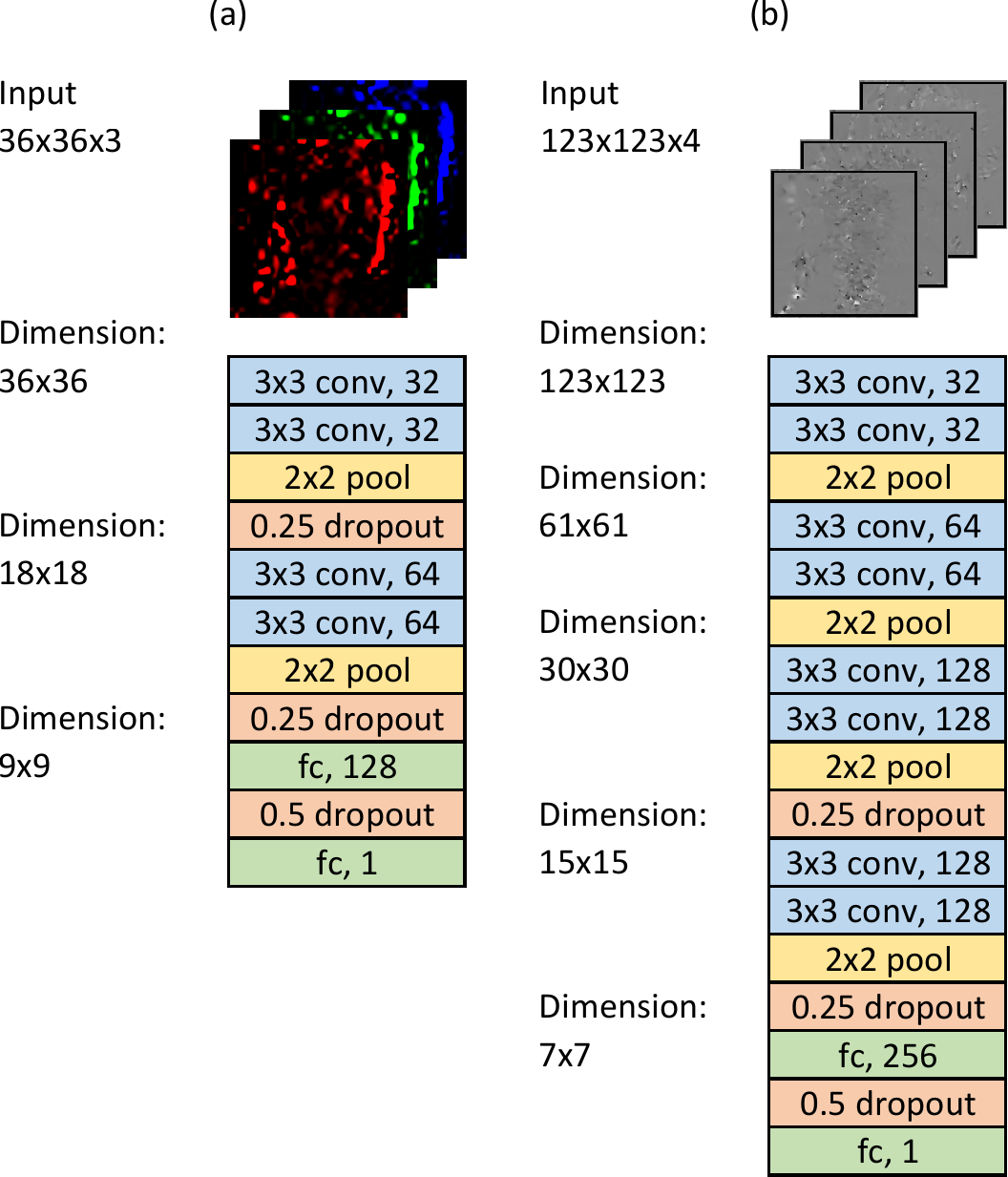}
	\caption{We used two exemplar tasks to illustrate the benefits of DeepMag. a) Color (Blood flow) magnification. b) Motion (respiration) magnification. These two tasks require different input motion representations and CNN architectures due to the nature of the motion signals.}
	\label{fig:cnn}
\end{figure}

\subsection{Example I: Color Magnification}
One example of applying our proposed algorithm is in the magnification of subtle skin color changes associated with the cardiac cycle. As blood flows through the skin it changes the light reflected from it. A good motion representation for these color changes is normalized frame difference \cite{chen2018deepphys}, which is summarized below.

For modeling lighting, imagers and physiology, previous works used the Lambert-Beer law (LBL) \cite{lam2015robust,Xu2014a} or Shafer's dichromatic reflection model (DRM) \cite{Wang2016b}. 
We build our motion representation on top of the DRM as it provides a better framework for separating specular reflection and diffuse reflection. Assume the light source has a constant spectral composition but varying intensity. We can define the RGB values of the $k$-th skin pixel in an image sequence by a time-varying function:
\begin{equation} \label{eq:1}
\pmb{C}_k(t)=I(t) \cdot (\pmb{v}_s(t)+\pmb{v}_d(t))+\pmb{v}_n(t)
\end{equation}
where $\pmb{C}_k(t)$ denotes a vector of the RGB values; $I(t)$ is the luminance intensity level, which changes with the light source as well as the distance between the light source, skin tissue and camera; $I(t)$ is modulated by two components in the DRM: specular reflection $\pmb{v}_s(t)$, mirror-like light reflection from the skin surface, and diffuse reflection $\pmb{v}_d(t)$, the absorption and scattering of light in skin-tissues; $\pmb{v}_n(t)$ denotes the quantization noise of the camera sensor.
$I(t)$, $\pmb{v}_s(t)$ and $\pmb{v}_d(t)$ can all be decomposed into a stationary and a time-dependent part through a linear transformation \cite{Wang2016b}:
\begin{equation} \label{eq:2}
\pmb{v}_d(t) = \pmb{u}_d \cdot d_0 + \pmb{u}_p \cdot p(t)  
\end{equation}
where $\pmb{u}_d$ denotes the unit color vector of the skin-tissue; $d_0$ denotes the stationary reflection strength; $\pmb{u}_p$ denotes the relative pulsatile strengths caused by hemoglobin and melanin absorption; $p(t)$ denotes the BVP.
\begin{equation} \label{eq:3}
\pmb{v}_s(t) = \pmb{u}_s \cdot (s_0+s(t)) 
\end{equation}
where $\pmb{u}_s$ denotes the unit color vector of the light source spectrum; $s_0$ and $s(t)$ denote the stationary and varying parts of specular reflections.  
\begin{equation} \label{eq:4}
I(t) = I_0 \cdot (1+i(t)) 
\end{equation}
where $I_0$ is the stationary part of the luminance intensity, and $I_0\cdot i(t)$ is the intensity variation observed by the camera.
The stationary components from the specular and diffuse reflections can be combined into a single component representing the stationary skin reflection:
\begin{equation} \label{eq:5}
\pmb{u}_c \cdot c_0 = \pmb{u}_s \cdot s_0 + \pmb{u}_d \cdot d_0
\end{equation}
where $\pmb{u}_c$ denotes the unit color vector of the skin reflection and $c_0$ denotes the reflection strength. Substituting (\ref{eq:2}), (\ref{eq:3}), (\ref{eq:4}) and (\ref{eq:5}) into (\ref{eq:1}), produces:
\begin{equation} \label{eq:6}
\pmb{C}_k(t)=I_0\cdot (1+i(t)) \cdot
(\pmb{u}_c \cdot c_0+\pmb{u}_s \cdot s(t)+\pmb{u}_p \cdot p(t))+\pmb{v}_n(t)
\end{equation}
As the time-varying components are much smaller (i.e., orders of magnitude) than the stationary components in (\ref{eq:6}), we can neglect any product between varying terms and approximate $\pmb{c}_k(t)$ as:
\begin{equation} \label{eq:7}
\pmb{C}_k(t)\approx \pmb{u}_c \cdot I_0 \cdot c_0 \cdot (1+i(t)) + 
\pmb{u}_s \cdot I_0 \cdot s(t)+\pmb{u}_p \cdot I_0 \cdot p(t)+\pmb{v}_n(t)
\end{equation}
The first step in computing our motion representation is
spatial averaging of pixels, which has been widely used for reducing the camera quantization error $\pmb{v}_n(t)$ in (\ref{eq:7}). We implemented this by downsampling every frame to $L$ pixels by $L$ pixels using bicubic interpolation. Emperical evidence shows that bicubic interpolation preserves the color information more accurately than linear interpolation~\cite{mcduff2018super}. Selecting $L$ is a trade-off between suppressing camera noise and retaining spatial resolution (\cite{wang2015exploiting} found that $L = 36$ was a good choice for face videos.) The downsampled pixel values will still obey the DRM model only without the camera quantization error:
\begin{equation} \label{eq:8}
\pmb{C}_l(t)\approx \pmb{u}_c \cdot I_0 \cdot c_0 + \pmb{u}_c \cdot I_0 \cdot c_0 \cdot i(t) + 
\pmb{u}_s \cdot I_0 \cdot s(t)+\pmb{u}_p \cdot I_0 \cdot p(t)
\end{equation}
where $l=1,\cdots,L^2$ is the new pixel index in every frame.

Then we need to reduce the dependency of $\pmb{C}_l(t)$ on the stationary skin reflection color $\pmb{u}_c \cdot I_0 \cdot c_0$, resulting from the light source and subject's skin tone. In (\ref{eq:8}), $\pmb{u}_c \cdot I_0 \cdot c_0$ appears twice. It is difficult to eliminate the second term as it interacts with the unknown $i(t)$. However, the first time-invariant term, which is usually dominant, can be removed by taking the first order derivative of both sides of (\ref{eq:8}) with respect to time:
\begin{equation} \label{eq:9}
\pmb{C}'_l(t)\approx \pmb{u}_c \cdot I_0 \cdot c_0 \cdot i'(t) +
\pmb{u}_s \cdot I_0 \cdot s'(t)+\pmb{u}_p \cdot I_0 \cdot p'(t)
\end{equation}

One problem with this frame difference representation is that the stationary luminance intensity level $I_0$ is spatially heterogeneous due to different distances to the light source and uneven skin contours. The spatial distribution of $I_0$ has nothing to do with physiology, but is different in every video recording setup. Thus, $\pmb{C}'_l(t)$ was normalized by dividing it by the temporal mean of $\pmb{C}_l(t)$ to remove $I_0$:
\begin{equation} \label{eq:10}
\frac{\pmb{C}'_l(t)}{\overline{\pmb{C}_l(t)}}\approx \pmb{1} \cdot i'(t) + diag^{-1}(\pmb{u}_c)\pmb{u}_s \cdot \frac{s'(t)}{c_0} + 
diag^{-1}(\pmb{u}_c)\pmb{u}_p \cdot \frac{p'(t)}{c_0}
\end{equation}
where $\pmb{1}=[1~1~1]^T$. In (\ref{eq:10}), $\overline{\pmb{C}_l(t)}$ needs to be computed pixel-by-pixel over a short time window to minimize occlusion problems and prevent the propagation of errors. We found it was feasible to compute it over two consecutive frames so that (\ref{eq:10}) can be expressed discretely as:
\begin{equation} \label{eq:11}
\pmb{X}_1(l,t) = \frac{\pmb{C}'_l(t)}{\overline{\pmb{C}_l(t)}}\sim\frac{\pmb{C}_l(t+1)-\pmb{C}_l(t)}{\pmb{C}_l(t+1)+\pmb{C}_l(t)}
\end{equation}
which is the normalized frame difference we used as motion representation.

The CNN we used for extracting pulse signals from the motion representation is shown in Fig. \ref{fig:cnn} (a). The pooling layers are 2x2 average pooling, and the convolution layers have a stride of one. All the layers use ReLU as the activation function. Note that bounded activation function such as tanh and sigmoid are not suitable for this task, as they will limit the extent to which the motion representation can be magnified in the gradient ascent.

After gradient ascent, the input motion representation $\pmb{X}_1(l,t)$ was magnified as $\pmb{X}_N(l,t)$, from which we could reconstruct the magnified video. The first step of reconstruction is to denoise the output motion representation by filtering the accumulated gradient:
\begin{equation}
\widetilde{\pmb{X}_N}(l,t) = \pmb{X}_1(l,t) + \mathcal{F}(\pmb{X}_N(l,t) - \pmb{X}_1(l,t))
\end{equation}
in which $\mathcal{F}$ is a zero-phase band-pass filter.  Note that unlike previous motion magnification methods the function of the filter here is not to select the target motion but to remove low and high frequency noise, so the filter bands do not need to precisely match the motion frequency in the video and can be chosen conservatively. Specifically, a 6th-order Butterworth filter with cut-off frequencies of 0.7 and 2.5 Hz was used to generally cover the normal heart rate range (42 to 150 beats per minute). Then we applied the inverse operation of (\ref{eq:11}) to reconstruct the downsampled version of the frames $\widetilde{\pmb{C}_l(t)}$:
\begin{equation}
\widetilde{\pmb{C}_l}(t+1) = \frac{1+\widetilde{\pmb{X}_N}(l,t)}{1-\widetilde{\pmb{X}_N}(l,t)} \cdot\widetilde{\pmb{C}_l}(t), ~\widetilde{\pmb{C}_l}(1)=\pmb{C}_l(1)
\end{equation}
Finally, $\pmb{C}_l(t)$ was upsampled back to the original video resolution:
\begin{equation}
\widetilde{\pmb{C}_k}(t) = \pmb{C}_k(t) - \mathcal{U}(\pmb{C}_l(t)) + \mathcal{U}(\widetilde{\pmb{C}_l}(t))
\end{equation}
in which $\mathcal{U}$ is an image upsampling operator.

\subsection{Example II: Motion Magnification}

Our second example is amplifying subtle motions on the human body induced by respiration. We used phase variations in a complex steerable pyramid to represent the local motions in a video. The complex steerable pyramid \cite{Simoncelli1992,Portilla2000} is a filter bank that breaks each frame of the video $C(t)$ into complex-valued sub-bands corresponding to different scales and orientations. The basis functions of this transformation are scaled and oriented Gabor-like wavelets with both cosine- and sine-phase components. Each pair of cosine- and sine-like filters can be used to separate the amplitude of local wavelets from their phase. Specifically, each scale $r$ and orientation $\theta$ is a complex image that can be expressed in terms of amplitude $A$ and phase $\phi$ as:
\begin{equation}
A(r,\theta,t)e^{i\phi(r,\theta,t)}
\end{equation}
We take the first-order temporal derivative of the local phases $\phi$ computed in this equation as our input motion representation:
\begin{equation} \label{eq:12}
X_1(r,\theta,t)=\phi(r,\theta,t+1)-\phi(r,\theta,t)
\end{equation}
For small motions, these phase variations are approximately proportional to displacements of image structures along the corresponding orientation and scale \cite{gautama2002phase}. To lower computational cost, we computed a pyramid with octave bandwidth and four orientations ($\theta=0^\circ,45^\circ,90^\circ,135^\circ$). Using half-octave or quarter-octave bandwidth and more orientations would enable our algorithm to amplify more motion details, but would require significantly greater computational recourses. In theory, $X_1(r,\theta,t)$ contains $r=1,2,\cdots,R$ scales of representations in different spatial resolutions, and extracting the target respiration motion from them would need $R$ different CNNs to fit different input dimensions. However, we found that $X_1(r,\theta,t)$ and the amplified $X_N(r,\theta,t)$ on different scales were approximately proportional to $0.5^r$, so it is possible to only process one scale $r=r_0$ and interpolate the other scales with it.

The CNN we used for extracting respiration signals from the motion representation is shown in Fig. \ref{fig:cnn} (b). The neural network is deeper than the one used for pulse magnification, because the input motion representation for respiration has a higher dimension. The pooling layers and convolution layers are of the same type as in Fig. \ref{fig:cnn} (a). As we met the dying ReLU problem (ReLU neurons were stuck in the negative side and always output 0) in our experiments, the activation functions of all the layers were replaced with scaled exponential linear units (SELU) \cite{klambauer2017self}.

After gradient ascent, the input motion representation $X_1(r_0,\theta,t)$ was magnified as $X_N(r_0,\theta,t)$, from which we could reconstruct the magnified video. Unlike in Example I, the phase variations were reconstructed by reversing (\ref{eq:12}) before denoising:
\begin{equation}
\widetilde{\phi}(r_0,\theta,t+1) = X_N(r_0,\theta,t)+\widetilde{\phi}(r_0,\theta,t),~
\widetilde{\phi}(r_0,\theta,1)=\phi(r_0,\theta,1)
\end{equation}
Then the reconstructed phase was denoised by band-pass filtering and $2\pi$ phase clipping:
\begin{equation}
\widetilde{\phi}(r_0,\theta,t) = \phi(r_0,\theta,t)
+\mathcal{F}(\widetilde{\phi}(r_0,\theta,t))\cdot\frac{sgn(2\pi-|\phi(r_0,\theta,t)|)+1}{2}
\end{equation}
The filter $\mathcal{F}$ is a 6th-order zero-phase Butterworth filter with cut-off frequencies of 0.16 and 0.5 Hz for generally covering the normal breathing rate range (10 to 30 beats per minute). The magnified phase of the other scales can be interpolated by exponentially scaling the filtered term:
\begin{equation} \label{eq:13}
\widetilde{\phi}(r_0,\theta,t) = \phi(r_0,\theta,t)
+\mathcal{F}(\widetilde{\phi}(r_0,\theta,t))\cdot\frac{sgn(2\pi-|\phi(r_0,\theta,t)|)+1}{2}\cdot(\frac{1}{2})^{r-r_0}
\end{equation}
Finally, the magnified video frame $\widetilde{C}(t)$ can be reconstructed from all the scales of the complex steerable pyramid with their phase updated as (\ref{eq:13}).

\begin{figure}[t]
	\centering\noindent
	\includegraphics[width=0.75\linewidth]{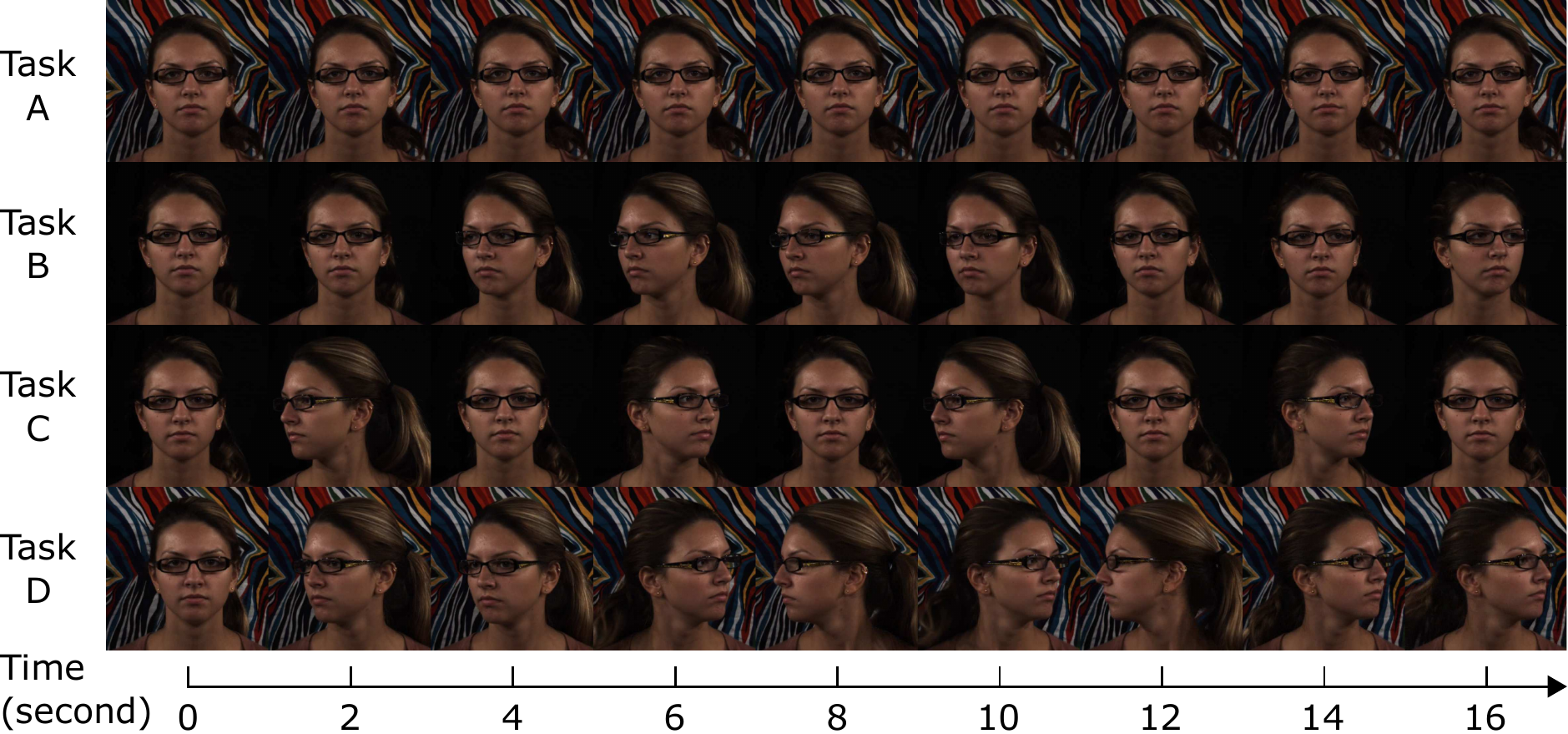}
	\caption{Exemplary frames from the four tasks of our video dataset. Note the different backgrounds and head rotation speeds.}
	\label{fig:frame_example}
\end{figure}

\section{Data}
We used the dataset collected by Estepp et al.~\cite{estepp2014recovering} for testing our approach.
Videos were recorded with a Basler Scout scA640-120gc GigE-standard, color camera, capturing 8-bit, 658x492 pixel images, 120 fps. The camera was equipped with 16 mm fixed focal length lens. Twenty-five participants (17 males) were recruited to participate for the study.
Nine individuals were wearing glasses, eight had facial hair, and four were wearing makeup on their face and/or neck.  The participants exhibited the following estimated Fitzpatrick Sun-Reactivity Skin Types~\cite{fitzpatrick1988validity}: I-1, II-13, III-10, IV-2, V-0.
Gold-standard physiological signals were measured using a BioSemi ActiveTwo research-grade biopotential acquisition unit.

We used videos of participants during a set of four, five-minutes tasks for our analysis. Two of the tasks (A and D) were performed in front of a patterned background and two (B and C) were performed in front of a black background. The four tasks were designed to capture different levels of head rotation about the vertical axis (yaw). Examples of frames from the tasks can be seen in Figs.~\ref{fig:frame_example}.
\newline
\textbf{Task A:} Participants stayed still allowing for small natural motions.
\newline
\textbf{Task B:} Participants performed a 120-degree sweep centered about the camera at a speed of 10 degrees/sec.
\newline
\textbf{Task C:} Similar to Task B but with a speed of 30 degrees/sec.
\newline
\textbf{Task D:} Participants were asked to reorient their head position once per second to a randomly chosen targets positioned in 20-degree increments over a 120-degree arc. Thus simulating random head motion.

\section{Evaluation}
We compare the color magnification results to Eulerian video magnification \cite{wu2012eulerian} and video acceleration magnification \cite{Zhang2017}, and compare the motion magnification results to phase-based Eulerian video magnification \cite{wadhwa2013phase} and video acceleration magnification (EVM and phase-based EVM perform poorly for motion magnification and color magnification respectively). In each case we perform qualitative evaluations similar to that presented in prior work. In addition, we perform a quantitative evaluation by assessing the image quality of the resulting videos. Prior work has generally not considered quantitative evaluations.

For obtaining our own results, the CNN model was either trained and tested on different time periods of the same videos (participant-dependent) or trained and tested on videos of different human participants (participant-independent), both using a 20\% holdout rate for testing. The qualitative and quantitative results we show in the following sections are always from video excerpts in the test set. To achieve a fair comparison, all the compared methods used the same filter bands: [0.7 Hz, 2.5 Hz] for pulse color magnification, and [0.16 Hz, 0.5 Hz] for respiration motion magnification. Since VAM uses difference of Gaussian (DoG) filters defined by a single pass-band frequency, we adopted the center frequencies of the physiology frequency bands ($\sqrt{0.7\times 2.5}=1.3~Hz$ for pulse, and $\sqrt{0.16\times 0.5}=0.28~Hz$ for respiration) as its filtering parameters. In the color magnification baselines, video frames were decomposed into multiple scales using a Gaussian pyramid with the intensity changes in the fourth level amplified (following the source code released by \cite{wu2012eulerian}). All the motion magnification baselines used complex steerable pyramids with octave bandwidth and four orientations. The magnification factors of all the methods were tuned to be visually the same on task A without head motion interferences.

\begin{figure}[!ht]
	\centering\noindent
	\includegraphics[width=0.72\linewidth]{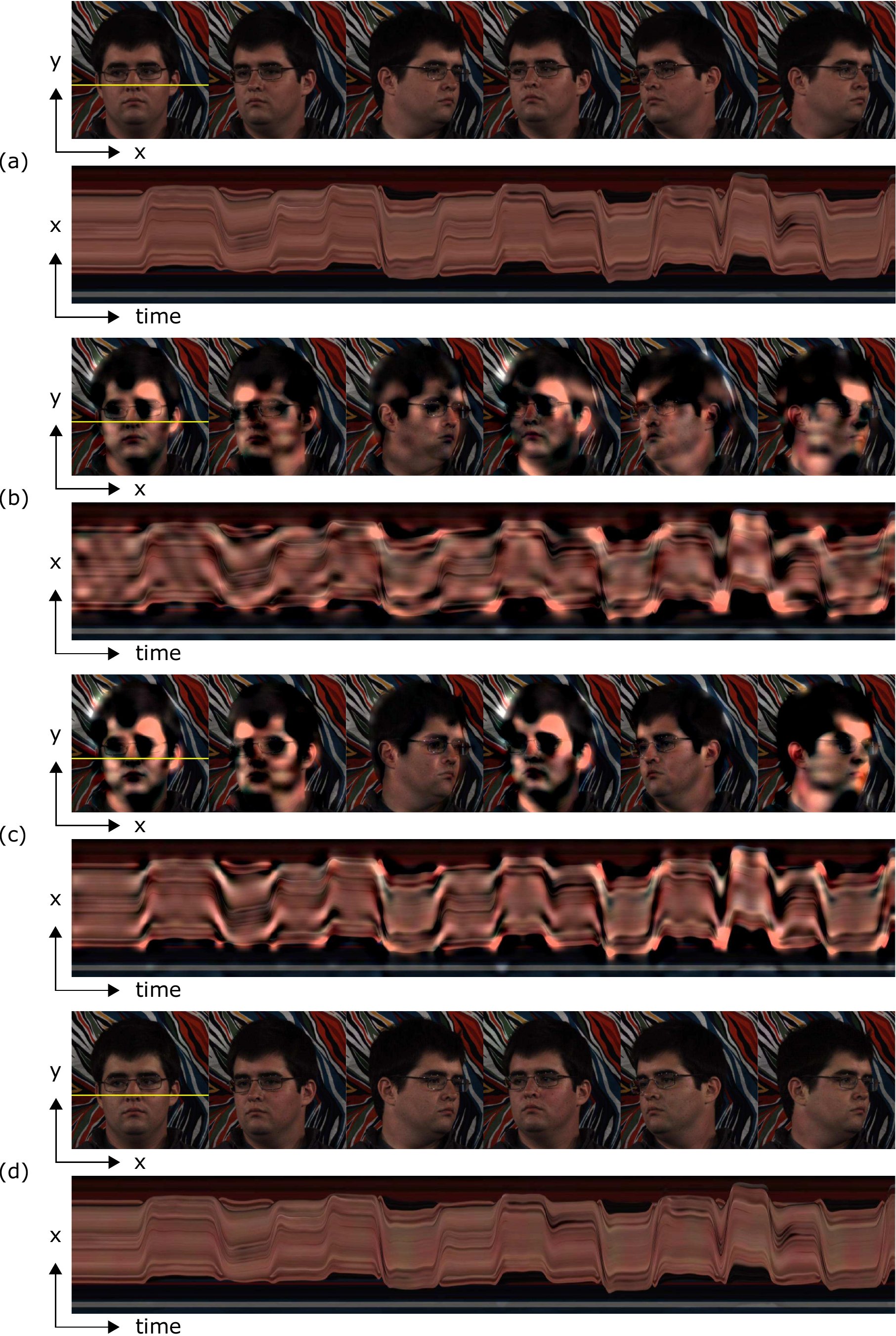}
	\caption{Scan line comparisons of color magnification methods for a Task D video: a) original video, b) Eulerian video magnification, c) video acceleration magnification, d) Our method. The yellow line shows the source of the scan line in the frames. The section of video shown was 15 seconds in duration. Our method produces clearer magnification of the color change due to blood flow and significantly fewer artifacts.}
	\label{fig:scanline_hr}
\end{figure}

\begin{figure}[!ht]
	\centering\noindent
	\includegraphics[width=0.65\linewidth]{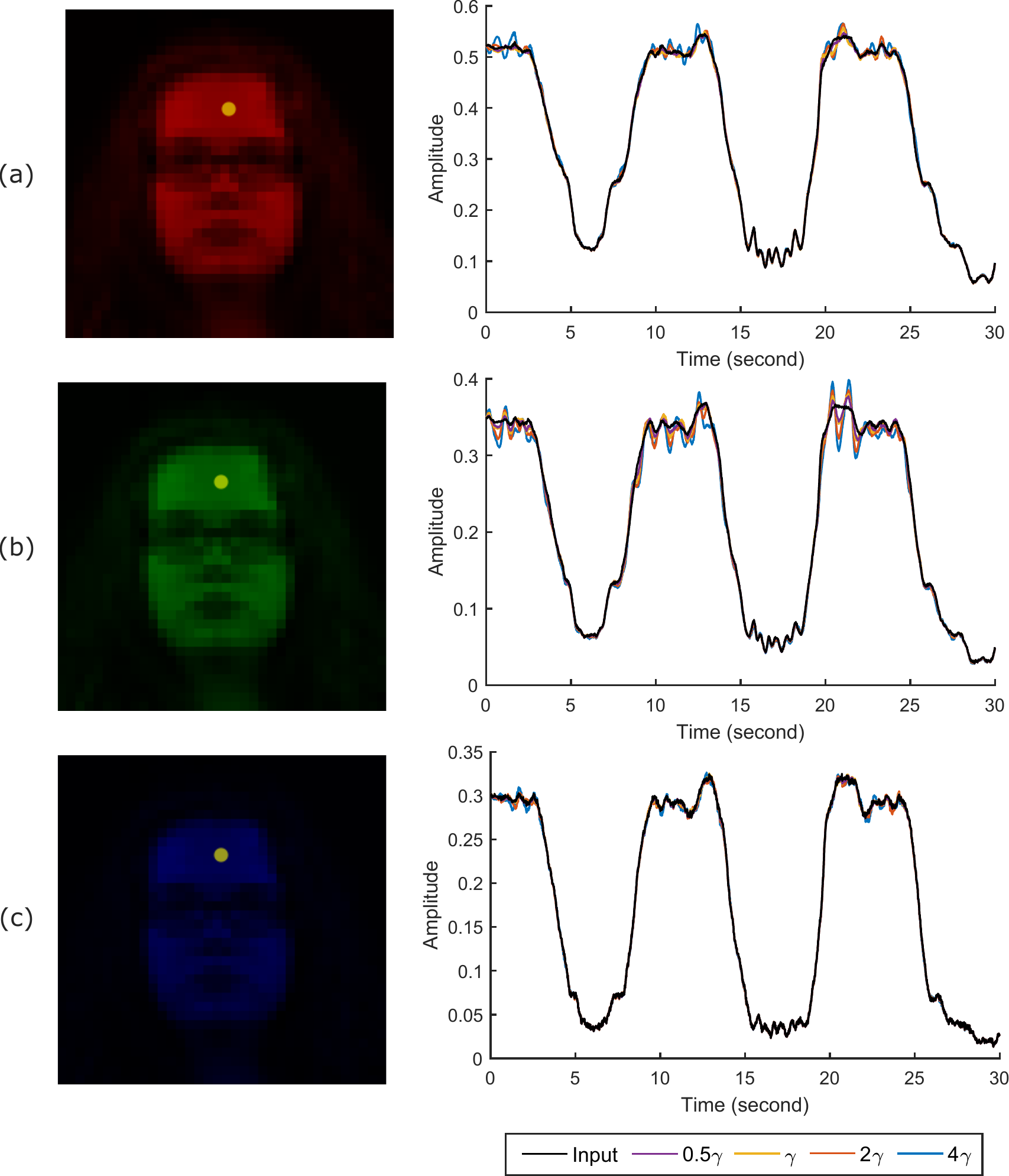}
	\caption{Original and magnified traces of a pixel (the yellow dot) in three color channels of a Task B video (a) red channel (b) green channel (c) blue channel. Magnified traces using different step sizes $\gamma$ are shown in different colors. The notches in the traces correspond to when the participant rotated her head to the far left/right and the pixel was no longer on the skin. Our method amplified the subtle color changes of the pixel only when it was on the skin, and kept the relative magnitudes of the pulse in three color channels with the green channel one being the strongest.}
	\label{fig:factor_hr}
\end{figure}

\subsection{Color Magnification}
We apply our method to the task of magnifying the photoplethysmogram. In this task the target variable for training the CNN was the gold standard contact PPG signal. The input motion representation was 36 pixels $\times$ 36 pixels $\times$ 3 color channels. In terms of the hyper-parameters of gradient ascent, the number of iterations $N$ was chosen to be 20, and the step size $\gamma$ was chosen to be $6\times 10^{-5}$. We found these choices provided a moderate magnification level, equivalent to the magnification using EVM. Different choices of these hyper-parameters will be discussed in the following sections.

Fig. \ref{fig:scanline_hr} shows a qualitative comparison between our method and the baseline methods. The human participant in the video reoriented his head once per second to a random direction. In the horizontal scan line of the input video, only the head rotation is visible and the subtle color changes of the skin corresponding to pulse cannot be seen with the unaided eye. In the results of the baseline methods, strong motion artifacts are introduced. This is because the complex head motion is not distinguishable from the pulse signal in the frequency domain, so it is amplified along with the pulse. Since the pulse-induced color changes are several orders of magnitude weaker than the head motion, they are completely buried by the motion artifacts in the amplified video. The VAM scan line (Fig. \ref{fig:scanline_hr} (c)) shows slightly fewer artifacts than the EVM scan line (Fig. \ref{fig:scanline_hr} (b)) as the head rotation was occasionally semi-linear. On the other hand, our algorithm uses a deep neural network to separate the pulse signal from the head motion, and uses gradient ascent to specifically amplify it. Consequently, its scan line (Fig. \ref{fig:scanline_hr} (d)) preserves the morphology of the head rotation while revealing the periodic color changes clearly on the skin. 

To show the magnification effects on different colors and different object surfaces, we drew the original and magnified traces of a pixel in three color channels of a video in Fig. \ref{fig:factor_hr}. The human participant in the video rotated her head left and right, so the selected pixel was on her forehead in half of time and was on the black background in the other half of time (corresponding to the notches in the traces). First, the pulse-induced color changes were only magnified when the pixel was on the skin surface, which proved the good spatial specificity of our algorithm. Second, the magnified pulse signal has much higher amplitude in the green channel than in the other channels. This is consistent with previous findings that the amplitude of the human pulse is approximately 0.33:0.77:0.53 in RGB channels under a halogen lamp \cite{de2013robust}, and verifies that our algorithm faithfully kept the original physiological property in magnification. 
Third, we changed the chosen step size $\gamma$ to its multiples ($0.5\gamma$, $2\gamma$ and $4\gamma$) with the number of iterations $N$ unaltered, and visualized the resulting pixel traces also in Fig. \ref{fig:factor_hr}. There is a clear trend that longer step sizes lead to higher amplitudes of the magnified pulse.

\begin{table*}
	\centering\noindent
	\scriptsize
	\setlength\tabcolsep{3pt}
	\caption{Video quality measured via Peak Signal-to-Noise Ratio (PSNR) and structural similarity (SSIM) for the magnified videos. The baselines for color magnification were EVM~\cite{wu2012eulerian} and VAM \cite{Zhang2017}, and for motion magnification were phase-EVM~\cite{wadhwa2013phase} and VAM. The table shows the average metrics among all videos within each task, while the bar charts also show the standard deviations as error bars. Our models (both participant-dependent and participant independent) produce videos with higher PSNR and SSIM compared to the baselines for all tasks. The benefit of our model is particularly strong for videos with greater levels of head rotation.}
	\label{tab:video_quality}
	\begin{tabular}{rcccc|cccc|cccc|cccc}
		\multicolumn{17}{c}{\includegraphics[width=0.14\textwidth]{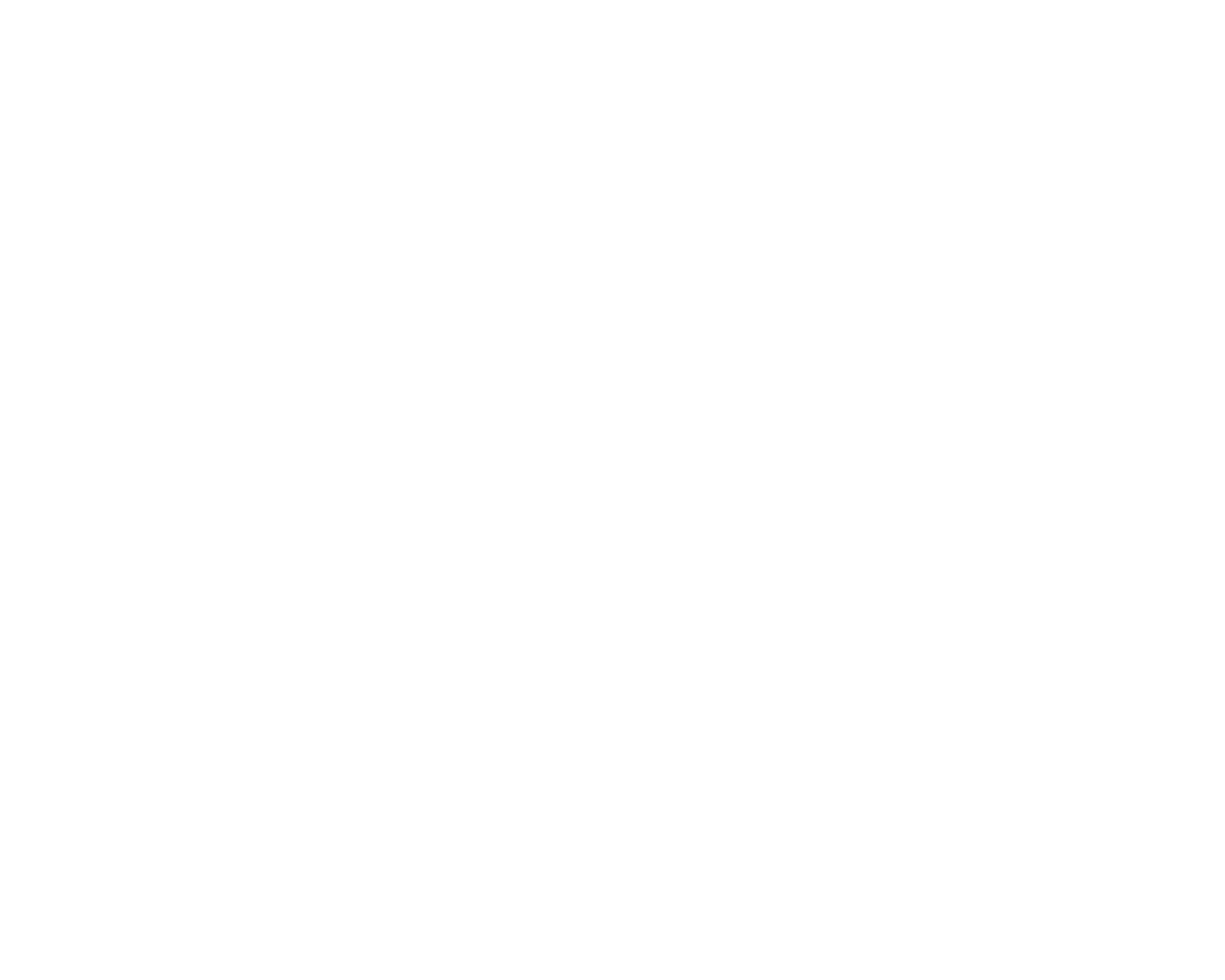} \includegraphics[width=0.82\textwidth]{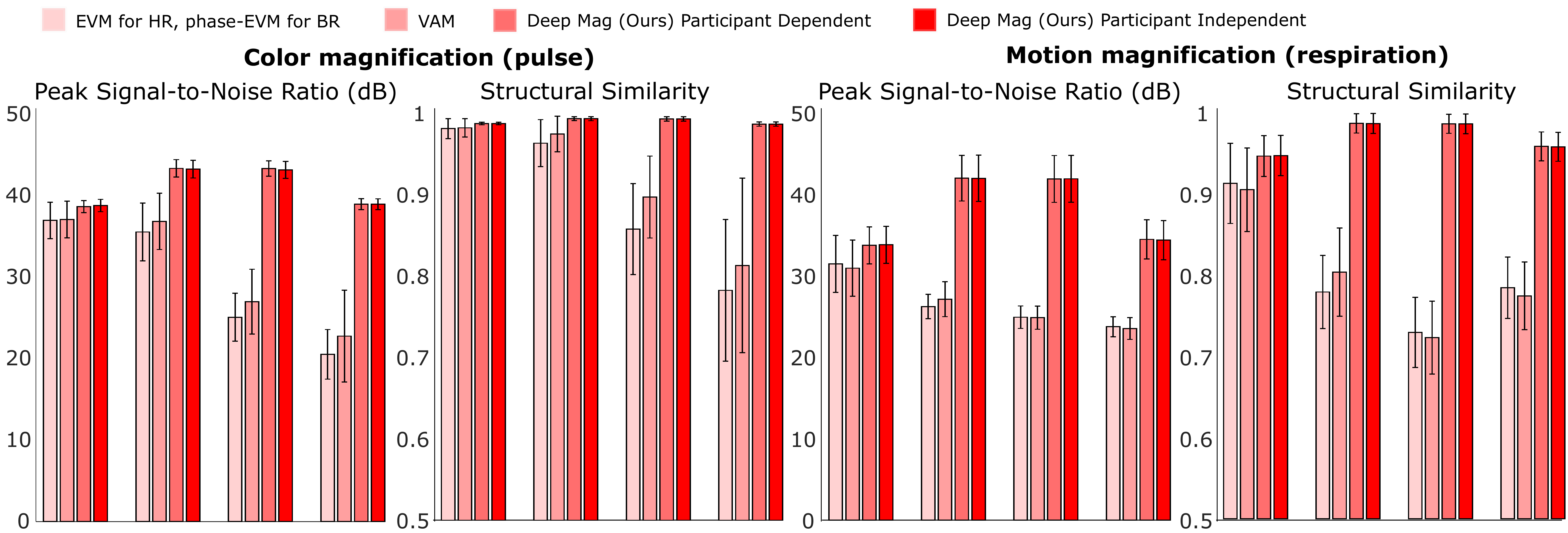}} \\
		Task & A & B & C & D & A & B & C & D & A & B & C & D & A & B & C & D \\
		\midrule
		EVM~\cite{wu2012eulerian} & 36.5 & 35.1 & 24.8 & 20.3 & .975 & .957 & .853 & .779 & - & - & - & - & - & - & - & - \\
		Phase-EVM~\cite{wadhwa2013phase} & - & - & - & - & - & - & - & - & 31.1 & 25.9 & 24.6 & 23.5 & .907 & .775 & .726 & .780 \\
		VAM~\cite{Zhang2017} & 36.6 & 36.4 & 26.7 & 22.5 & .976 & .969 & .892 & .809 & 30.6 & 26.8 & 24.6 & 23.2 & .900 & .800 & .720 & .770 \\
		DeepMag - P. Dep.  & 38.2 & \textbf{42.8} & \textbf{42.8} & \textbf{38.5} & \textbf{.981} & \textbf{.987} & \textbf{.987} & \textbf{.981} & 33.3 & \textbf{41.5} & 41.4 & \textbf{34.1} & \textbf{.940} & \textbf{.980} & \textbf{.979} & \textbf{.952} \\
		DeepMag - P. Ind. & \textbf{38.3} & 42.7 & 42.6 & \textbf{38.5} & \textbf{.981} & \textbf{.987} & \textbf{.987} & \textbf{.981} & \textbf{33.4} & \textbf{41.5} & \textbf{41.4} & 34.0 & \textbf{.940} & .979 & \textbf{.979} & .951 \\ 
		\bottomrule
	\end{tabular}
\end{table*}

To perform a quantitative evaluation of video quality we used two metrics: peak signal-to-noise ratio (PSNR) and structural similarity (SSIM). In both cases we calculated the metrics on every frame of the tested videos, and took their averages across all participants within each task. The reference frame in each case was the corresponding frame from the original, unmagnified video. Table~\ref{tab:video_quality} shows a comparison of the video quality metrics for the baselines and our method. 
Although the magnified blood flow or respiration will naturally cause the metrics to be lower, we found that artifacts in the generated videos had a much more significant impact on their values than the magnified physiology. Thus, lower PSNR and SSIM values indicate more artifacts and lower quality.
According to the table, our methods achieve both higher PSNR and SSIM than the baseline methods, which verify the ability of our methods to magnify subtle color changes with motion artifact suppressed. On task A containing limited head motions, the metrics of the baseline methods are very close to those of our method. However, as the head rotation becomes faster and random on more difficult tasks, the video quality of the baseline outputs dramatically decreases. This is because their algorithms amplify any motion lying in the filter band and does so indiscriminately. The magnification thus leads to significant artifact when large head motions are present. On the other hand using our method, the video quality is maintained at almost the same level on different tasks. Both PSNR and SSIM are only slightly lower on Task A and Task D, because the patterned background is more vulnerable to artifacts than the black one. The difference between the participant-dependent results and the participant-independent results is also very small, suggesting that our algorithm has good generalization ability and can be successfully applied to new videos containing different human participants without additional tuning. 

\begin{figure}[!ht]
	\centering\noindent
	\includegraphics[width=0.72\linewidth]{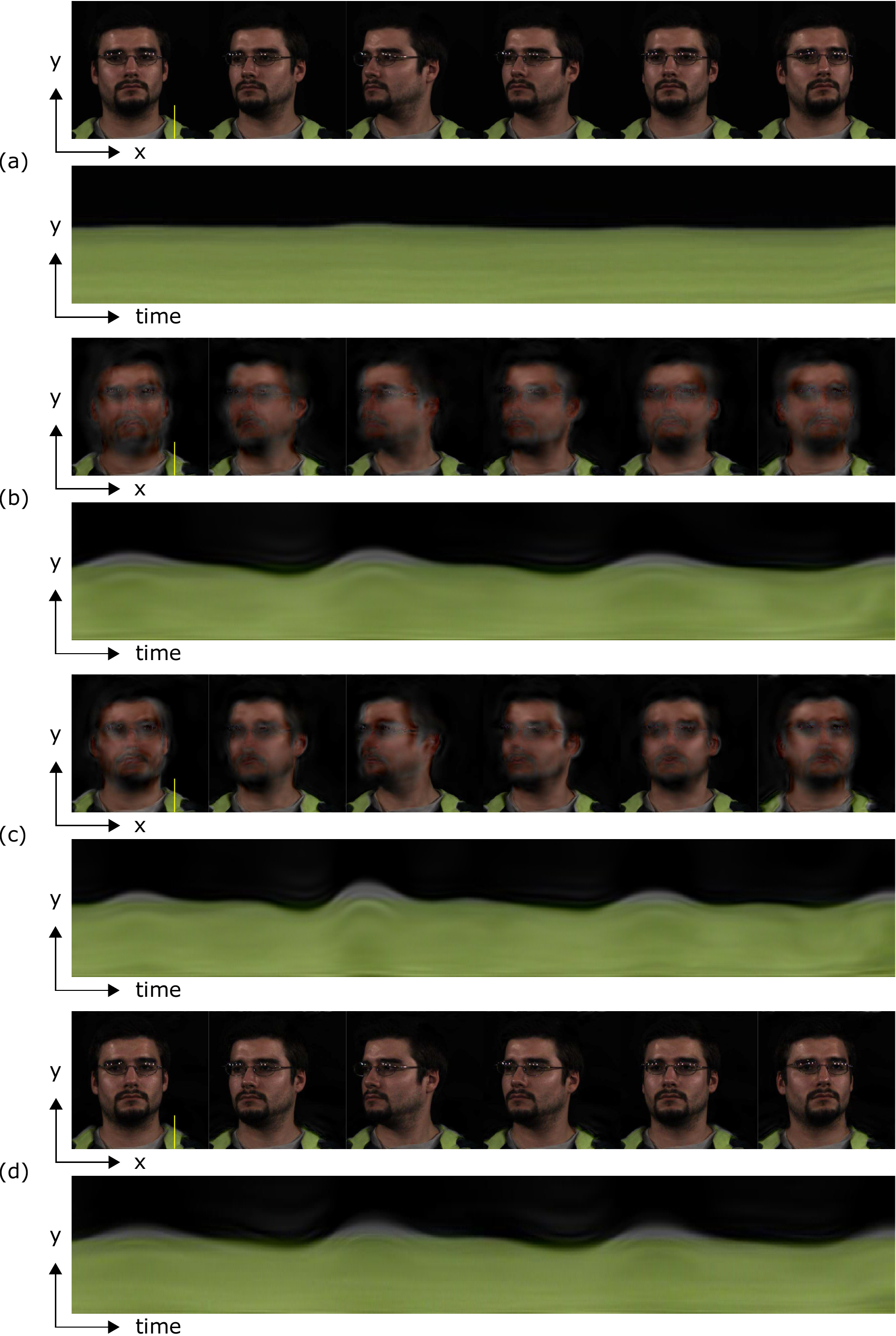}
	\caption{Scan line comparisons of motion magnification methods for a Task B video: a) original video, b) phase-based Eulerian video magnification, c) video acceleration magnification, d) Our method. The yellow line shows the source of the scan line in the frames. The section of video shown was 15 seconds in duration. Our method produces comparable magnification of the respiration motion and significantly fewer artifacts and blurring.}
	\label{fig:scanline_br}
\end{figure}

\begin{figure}[!ht]
	\centering\noindent
	\includegraphics[width=0.65\linewidth]{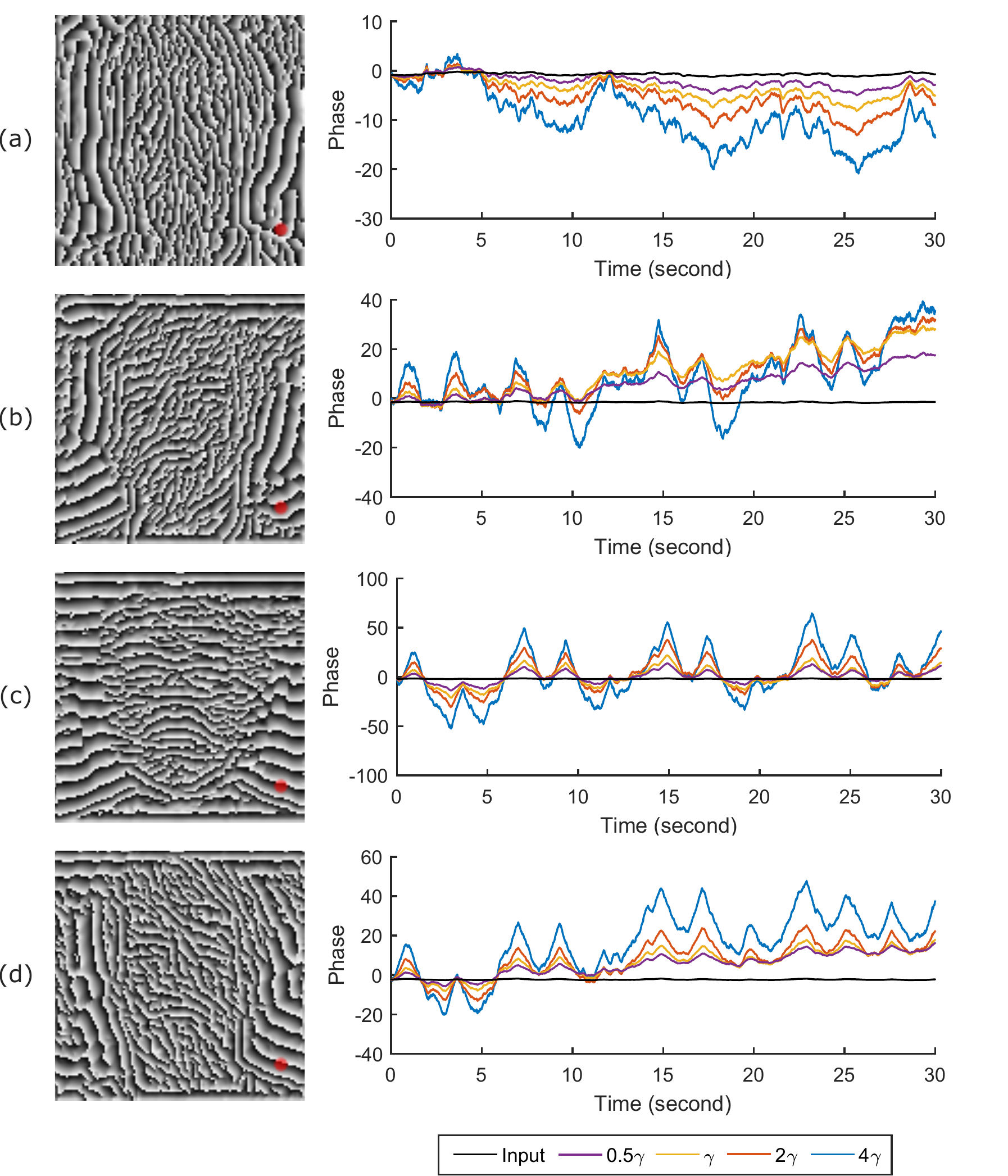}
	\caption{Original and magnified traces of a pixel (the red dot) in the phase representation $\phi(r_0,\theta,t)$ of a Task C video along four orientations (a) $\theta=0^{\circ}$ (b) $\theta=45^{\circ}$ (c) $\theta=90^{\circ}$ (d) $\theta=135^{\circ}$. Magnified traces using different step sizes $\gamma$ are shown in different colors. The pixel exhibits a respiration movement mainly in the vertical direction, so its magnified phase traces have the highest amplitude along the $\theta=90^{\circ}$ orientation.}
	\label{fig:factor_br}
\end{figure}

\subsection{Motion Magnification}

We apply our method to the task of magnifying respiration motions. In this task the target variable for training the CNN was the gold standard respiration signal measured via the chest strap. Given the subtle nature of the motions we found that a higher dimension input motion representation was needed than for the PPG magnification. As shown in Fig. \ref{fig:cnn}, the motion representation was in 123 pixels $\times$ 123 pixels $\times$ 4 orientations. The gradient ascent hyper-parameters $N$ and $\gamma$ were chosen to be 20 and $3.6\times 10^{-3}$ to produce moderate magnification effects.

Fig. \ref{fig:scanline_br} shows a qualitative comparison between our method and the baseline methods. The human participant in the video rotated his head at a speed of 10 degrees/sec. A vertical scanline on his shoulder was drawn along with time to show the respiration movement. In the input video, the respiration movement is very subtle. Both our method and the baseline methods greatly increased its magnitude (Fig. \ref{fig:scanline_br} (b) (c) (d)). However, the baseline methods cannot clearly distinguish the phase variations caused by respiration and by head rotation, so it also amplified the head rotation and blurred the participant's face. Our method is based on a better motion discriminator learned via the CNN so that the head motions are not amplified. 

To show the intermediate phase variations and different magnification effects along different orientations, we drew the original and magnified traces of a pixel in the phase representation $\phi(r_0,\theta,t)$ (Fig. \ref{fig:factor_br}). Since the selected pixel is on the shoulder of the human participant, the respiration movement is mainly in the vertical direction. As a result, the amplified phase variations corresponding to breathing have the highest amplitude along $\theta=90^{\circ}$ (Fig. \ref{fig:factor_br} (c)) and the lowest amplitude along $\theta=0^{\circ}$ (Fig. \ref{fig:factor_br} (a)). 
We also changed the chosen step size $\gamma$ to its multiples ($0.5\gamma$, $2\gamma$ and $4\gamma$) with the number of iterations $N$ unaltered, and visualized the resulting phase traces in Fig. \ref{fig:factor_br}. The figure suggests that the magnification level always increases along with the step size.

The same quantitative metrics as those for color magnification were computed and shown in  Table~\ref{tab:video_quality}. They also generally follow the same pattern as in the color magnification analysis: The video quality of the baseline methods is impacted by the level of head motions, while our method is considerably more robust. There is no significant difference between our participant-dependent results and participant-independent results.


\begin{table}
	\centering
	\small
	\caption{Video quality measured via Peak Signal-to-Noise Ratio (PSNR) and Structural Similarity (SSIM) for Task C videos magnified to different levels.}
	\label{tab:mag_factor}
	\begin{tabular}{rcccc|cccc}
		\toprule
		& \multicolumn{4}{c}{\textbf{PSNR (dB)}}  & \multicolumn{4}{c}{\textbf{SSIM}} \\
		Step size & $0.5\gamma$ & $\gamma$ & $2\gamma$ & $4\gamma$ & $0.5\gamma$ & $\gamma$ & $2\gamma$ & $4\gamma$ \\
		\midrule
		Pulse & 43.2 & 42.6 & 41.6 & 39.9 & 0.987 & 0.987 & 0.986 & 0.986 \\
		Respiration & 42.0 & 41.4 & 40.6 & 39.6 & 0.982 & 0.979 & 0.974 & 0.965 \\
		\bottomrule
	\end{tabular}
\end{table}

\begin{figure}[!ht]
	\centering\noindent
	\includegraphics[width=0.8\linewidth]{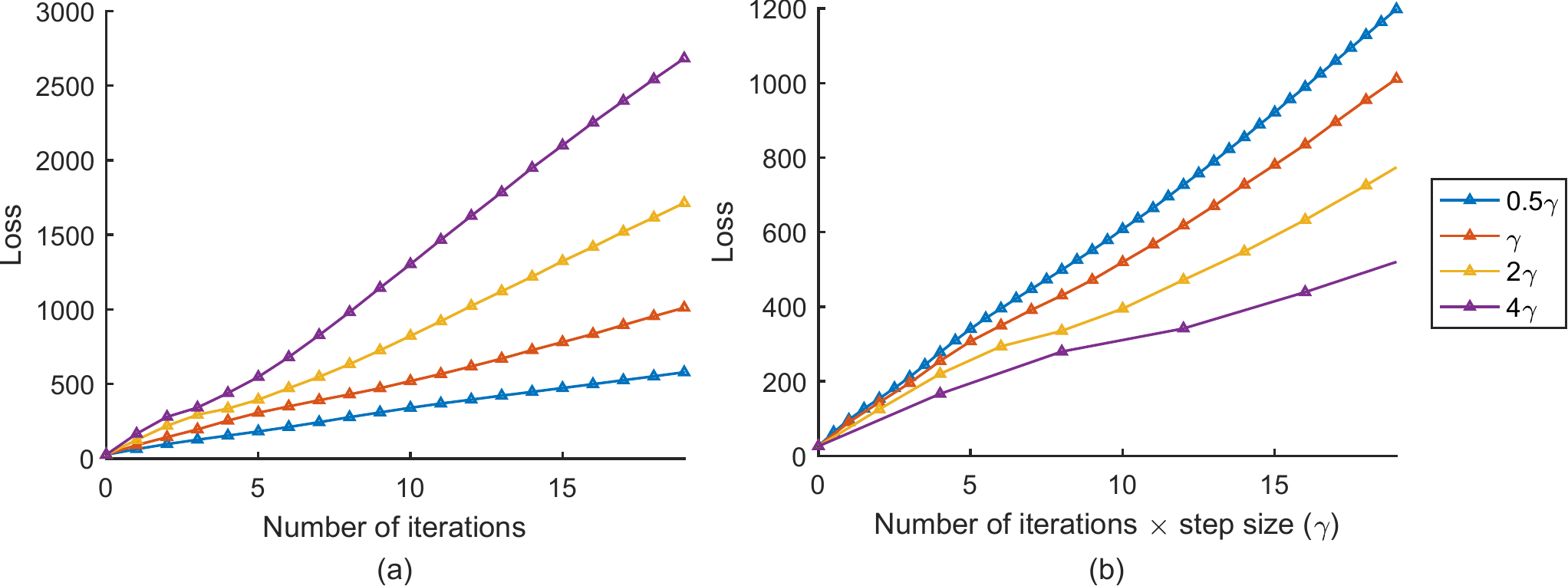}
	\caption{Learning curves: (a) The change of the CNN loss with different numbers of iterations $N$ and different step sizes $\gamma$. (b) The change of the CNN loss with different products of $N$ and $\gamma$.}
	\label{fig:learningcurve}
\end{figure}

\subsection{Magnification Factors}

The magnification factor of our algorithm is controlled by two hyper-parameters, the number of iterations $N$ and the step size $\gamma$. In Fig. \ref{fig:factor_hr} and Fig. \ref{fig:factor_br}, we chose the same $N$ and tuned $\gamma$ to be different multiples. The resulting magnification levels were always higher when $\gamma$ was longer. However, there is a trade-off in the selection of $\gamma$, as a higher magnification factor also introduces more artifacts. Table \ref{tab:mag_factor} shows the average video quality metrics PSNR and SSIM for our output videos on an exemplary task (Task C) with different choices of $\gamma$. For both the pulse and respiration magnification tasks, the video quality decreases to different extents with the increase of $\gamma$. Given that artifacts considerably reduce the PSNR and SSIM metrics (as shown in Table~\ref{tab:video_quality}), the fact that the values do not change dramatically with $\gamma$ shows that few artifacts are introduced with increasing magnification.

To quantitatively analyze the effects of $N$ and $\gamma$ on the magnification factor, we drew exemplary learning curves for one of our videos in Fig. \ref{fig:learningcurve} (a) with different choices of parameters. The curves show the changes of our CNN loss, the L2 norm of the differential motion signal, which is a good estimate of the target motion magnitude. According to the learning curves, both $N$ and $\gamma$ positively correlate with the motion magnitude, and the relationship between $N$ and the motion magnitude is semi-linear. However, a longer step size with fewer iterations is not equivalent to a shorter step with more iterations. In Fig. \ref{fig:learningcurve} (b), we show how the loss changes along with the product of $N$ and $\gamma$, which suggests that relatively small step sizes and more iterations can increase the magnification factor more efficiently.

\begin{figure}[!ht]
	\centering\noindent
	\includegraphics[width=0.8\linewidth]{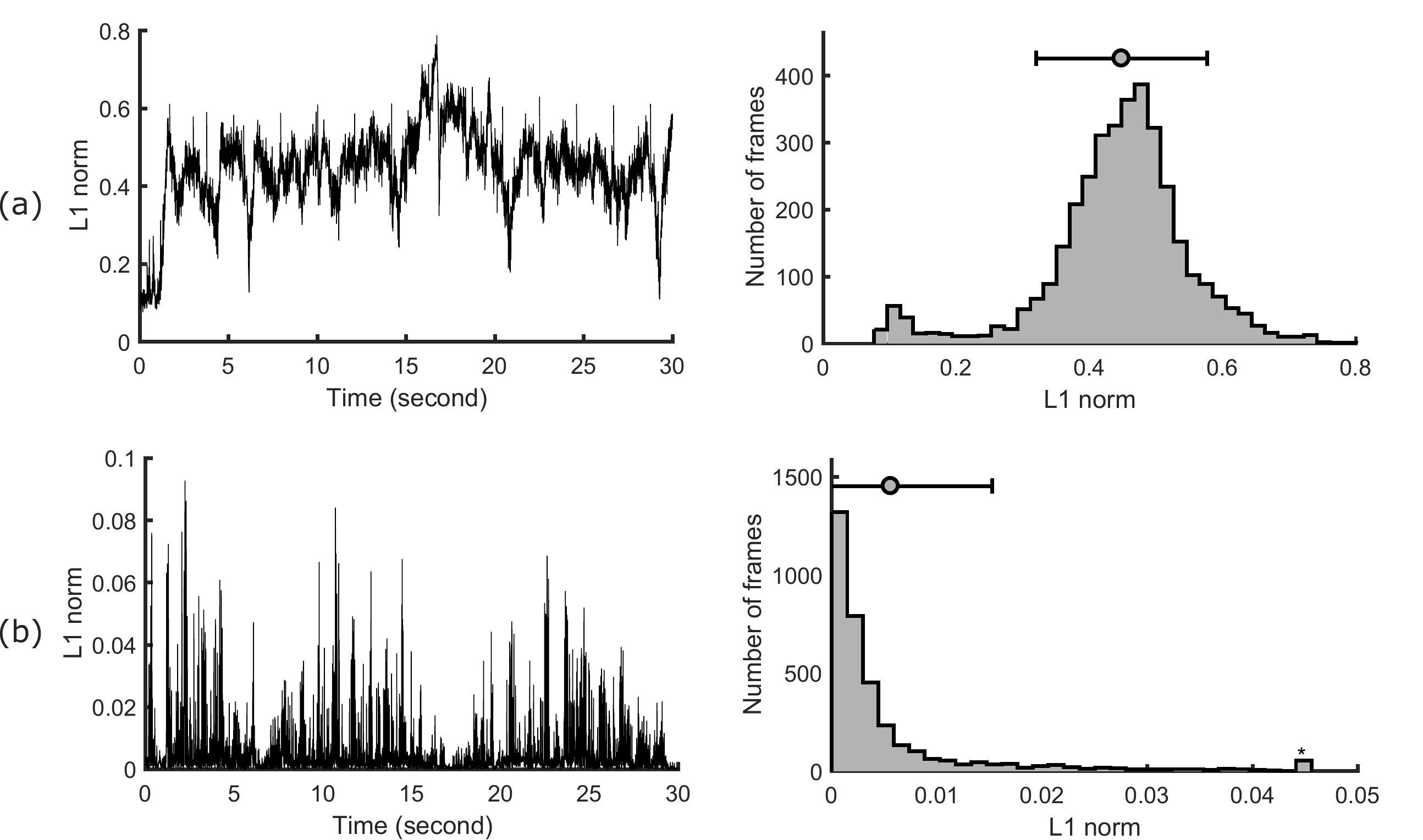}
	\caption{(a) Time series and histograms of the L1 norms of the input motion representation $X_{1}$ for a 30-second video. (b) Time series and histograms of the L1 norms of the motion gradient $\nabla\|y(X_{1}|\theta)\|_2$ for the same video.}
	\label{fig:comp_l1}
\end{figure}

\begin{figure}[!ht]
	\centering\noindent
	\includegraphics[width=0.8\linewidth]{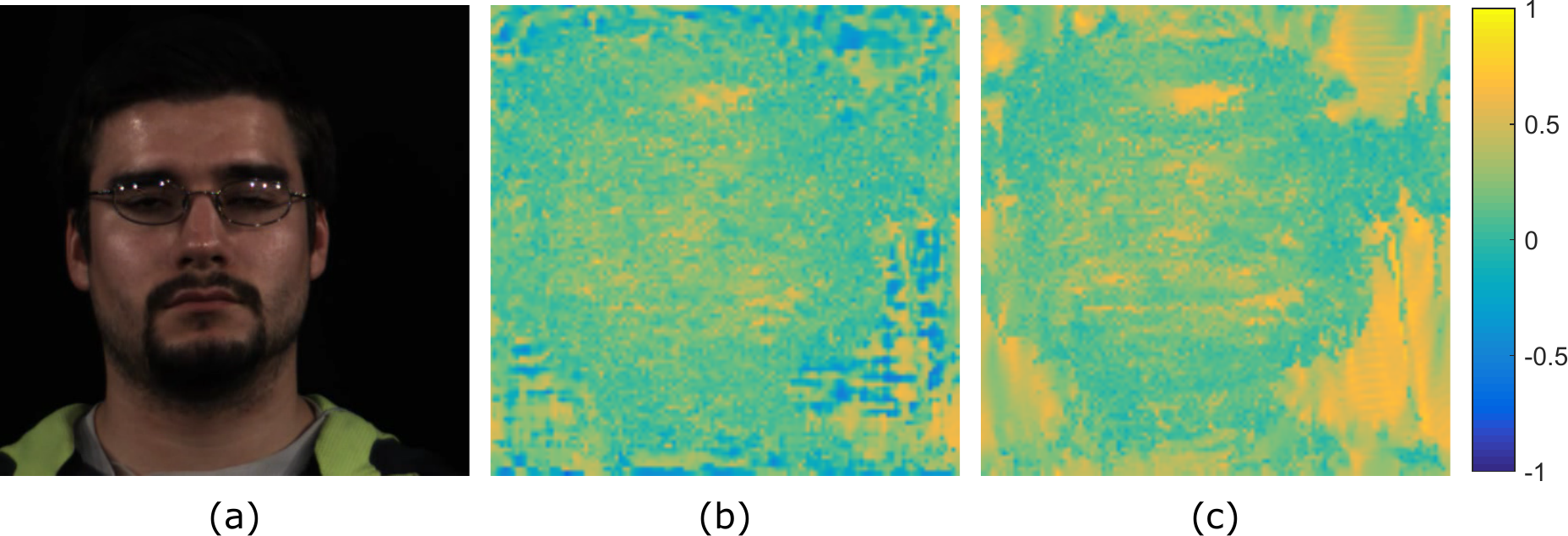}
	\caption{Pixel-wise correlation coefficients between the input and magnified motion representations in the respiration magnification task, with the sign correction mechanism (b) and without the sign correction mechanism (c).}
	\label{fig:flip}
\end{figure}

\subsection{Gradient Ascent Mechanisms}
Compared with traditional gradient ascent, we added two new mechanisms to adapt the approach to the task of video magnification: L1 normalization and sign correction. Here we show experimental results to support the necessity of these mechanisms.

The goal of applying L1 normalization is to make sure every frame in a video is magnified to the same level. To achieve this goal, the gradient $\nabla\|y(X_{n}|\theta)\|_2$ in (\ref{eq:I}) needs to be approximately proportional to the motion representation $X_{n}$. However, it was not the case without L1 normalization.  Fig. \ref{fig:comp_l1} shows the time series and histograms of the L1 norms of $X_{1}$ and $\nabla\|y(X_{1}|\theta)\|_2$ for a 30-second video. It is obvious that the distribution of the motion representation is Gaussian while the distribution of the gradient is highly skewed. To correct the distribution of the gradient to match the motion representation, it needs to be L1 normalized.

In Fig. \ref{fig:flip}, we show the pixel-wise correlation coefficients between the input and the magnified motion representations in the respiration magnification task, with and without the sign correction mechanism. When there is no sign correction, the correlation coefficients have both positive and negative values (Fig. \ref{fig:flip} (b)). As introduced in Section 3.1, the negative values appear because the target motion could be amplified with its direction reversed. In the example in Fig. \ref{fig:flip} (b), most of the negative values happen on the background, which are negligible as the background has nearly no motion to amplify, but some of them are on the human body, which will cause the output video to be blurry on magnification. After sign correction is applied, all the correlation coefficients become positive (Fig. \ref{fig:flip} (c)).

\section{Conclusions}
Revealing subtle signals in our everyday world is important for helping us understand the processes that cause them. 
We present a novel single deep neural framework for video magnification that is robust to large rigid motions. Our method leverages a CNN architecture that enables magnification of a specific source signal even if it overlaps with other motion sources in the frequency domain.  We present several methodological innovations in order to achieve our results, including adding L1 normalization and sign correction to the gradient ascent method.

Pulse and respiration magnification are good exemplar tasks for video magnification as these physiological phenomena cause both subtle color and motion variations that are invisible to the unaided eye. 
Our qualitative evaluation illustrates how the PPG color changes and respiration motions can be clearly magnified. Comparisons with baseline methods show that our proposed architecture dramatically reduces artifacts when there are other rotational head motions present in the videos. 

In a systematic quantitative evaluation our method improves the PSNR and SSIM metrics across tasks with different levels of rigid motion. By magnifying a specific source signal we are able to maintain the quality of the magnified videos to a greater extent.  

\printbibliography

\end{document}